\documentclass[conference]{IEEEtran}
\usepackage{cite}
\usepackage{amsmath}
\usepackage{array}
\usepackage{url}
\usepackage{graphicx}
\usepackage{booktabs}
\usepackage[table,xcdraw]{xcolor}
\usepackage{multirow}
\usepackage{balance}
\usepackage{fancyhdr}
\usepackage{hyperref}
\hypersetup{
     colorlinks   = true,
     allcolors    = black
}

\fancypagestyle{firststyle}
{
   \fancyhead[C]{\footnotesize 
           First published in the Proceedings of the 25th European Signal
           Processing Conference (EUSIPCO-2017) in 2017, published by EURASIP.
   }
}

\begin{document}
\title{End-to-End Musical Key Estimation Using a Convolutional Neural Network}

\author{\IEEEauthorblockN{Filip Korzeniowski and Gerhard Widmer}
\IEEEauthorblockA{Department of Computational Perception,\\
Johannes Kepler University, Linz, Austria\\
Email: filip.korzeniowski@jku.at}
}

\maketitle

\begin{abstract}
We present an end-to-end system for musical key estimation, based on a
convolutional neural network. The proposed system not only out-performs
existing key estimation methods proposed in the academic literature; it is also
capable of learning a unified model for diverse musical genres that performs
comparably to existing systems specialised for specific genres. Our experiments
confirm that different genres do differ in their interpretation of tonality,
and thus a system tuned e.g.\ for pop music performs subpar on pieces of
electronic music. They also reveal that such cross-genre setups evoke specific
types of error (predicting the relative or parallel minor). However, using the
data-driven approach proposed in this paper, we can train models that deal with
multiple musical styles adequately, and without major losses in accuracy.
\end{abstract}

\IEEEpeerreviewmaketitle
\thispagestyle{firststyle}

\section{Introduction}

The key of a piece of (Western) music defines its harmonic center and thus
plays a vital role in the piece's tonality. It gives meaning to the harmonic
progression of a piece, and provides the backdrop for the build-up and release
of harmonic tension. Thus, if we want to automatically detect tension and its
release in musical audio, if we want to find similarities in harmonic structure
in recordings of songs, or if we want to assist disc jockeys or creators of
electronic music in finding appropriate musical samples to work with, we need
reliable methods that extract the key from a musical piece. It is therefore
of no surprise that the automatic key estimation is a long-standing task in the
music information retrieval community.

Many key estimation systems proposed in the academic literature
rely on the same fundamental pipeline. First, a time-frequency
representation, like the spectrogram or the constant-q-transform, is computed
from the audio. Then, a variant of a pitch-class-profile, also called chroma
feature, is extracted for every time frame. Here, the goal is to get an
octave-independent representation of pitch classes that is robust to timbre.
Finally, these features are aggregated over time and matched with feature
templates for each key, where the key with the best matching template is
selected.  Examples of such systems are
\cite{pauws_musical_2004,temperley_what_1999,noland_signal_2007}, and, most
recently, \cite{faraldo_key_2016}. Such systems typically report a single
\emph{global key} for a piece. This limitation is reasonable for a variety of
genres (like pop/rock or electronic music), but fails to cope with pieces that
contain key modulations (as common in classical music).

Another line of work considers estimating key and chords simultaneously,
e.g.\ \cite{mauch_simultaneous_2010,pauwels_combining_2014,ni_endtoend_2012,digiorgi_automatic_2013}.
Such systems aim at exploiting the musicological relationship between key and
chords in order to improve the accuracy of both. They typically estimate
\emph{local keys}, and are thus able to cope with key modulations. However,
while they bear the potential to explain the harmonic content of musical audio
more holistically, dedicated systems currently seem to achieve better
results\footnote{See results of the yearly MIREX challenges at
  \url{www.music-ir.org/mirex}.}.

This paper targets the estimation of a \emph{single global key} for pieces of
musical audio. In contrast to previous works, we abandon hand-crafting or tuning
elements in the key estimation pipeline. Instead, we employ a convolutional
neural network that encompasses the three stages of pre-processing, feature
extraction, and classification. Although neural networks have been used for key
detection (e.g.\, \cite{sun_key_2009,dieleman_audiobased_2011}), the existing
approaches rely on a hand-crafted feature extraction stage: in
\cite{sun_key_2009}, a pitch class distribution matrix is fed into a neural
network classifier; in \cite{dieleman_audiobased_2011}, beat-aligned chroma and
timbre features are used as input. Our system operates directly on the
spectrogram, and it can estimate all its parameters from the data. To our
knowledge, this is the first work that replaces the complete key estimation
pipeline with a model that can be optimised in an end-to-end
manner\footnote{One might argue that \cite{ni_endtoend_2012} is also an
        ``end-to-end'' system that detects both chords and keys. However, they
        used feature extraction methods heavily based on expert knowledge
        (tuning correction, harmonic-percussive source separation, beat
        synchronisation, frequency-split chroma computation), and trained a
        dynamic Bayesian network whose structure is based on domain knowledge.
        Also, its key estimation performance was never evaluated.}.

\section{Method}

Our system consists of two steps: first, we compute from the audio a
logarithmically filtered log-magnitude spectrogram. This process is detailed in
Sec.~\ref{sec:input_processing}. Then, we feed this time-frequency
representation to the convolutional neural network, described in
Sec.~\ref{sec:model}, for classification.

\subsection{Input Processing}\label{sec:input_processing}

We input a spectral representation of the audio to our models. Based on our
previous work on extracting harmonic information from audio
\cite{korzeniowski_fully_2016,korzeniowski_feature_2016}, we first compute from
the audio the magnitude spectrogram $|\mathbf{S}|$ (frame size of 8192 at 5
frames per second, where the sample rate is 44.1 kHz); then, we apply a filterbank
$\mathbf{B}^{\triangle}_{\text{Log}}$ with logarithmically spaced triangular
filters (24 bands per octave, from 65 Hz to 2100 Hz), which results in a
time-frequency representation in which the fundamental frequencies of notes are
spaced linearly; finally, we logarithmise the magnitudes of the
filtered spectrogram to compress the value range, resulting in the
logarithmically filtered log-magnitude spectrogram

\[ \mathbf{L} = \log\left(1 + \mathbf{B}^{\triangle}_{\text{Log}} \left\lvert \mathbf{S}\right\rvert \right). \]

This representation is similar to a constant-q transform, but is much
cheaper to compute. Additionally, as shown in \cite{kelz_potential_2016}, the
constant-q transform does not necessarily lead to better results in tasks
relying on pitch information. All computations are done using the madmom
library \cite{bock_madmom_2016}.

\subsection{Model}\label{sec:model}

\begin{figure}
\centering
\includegraphics[width=\columnwidth]{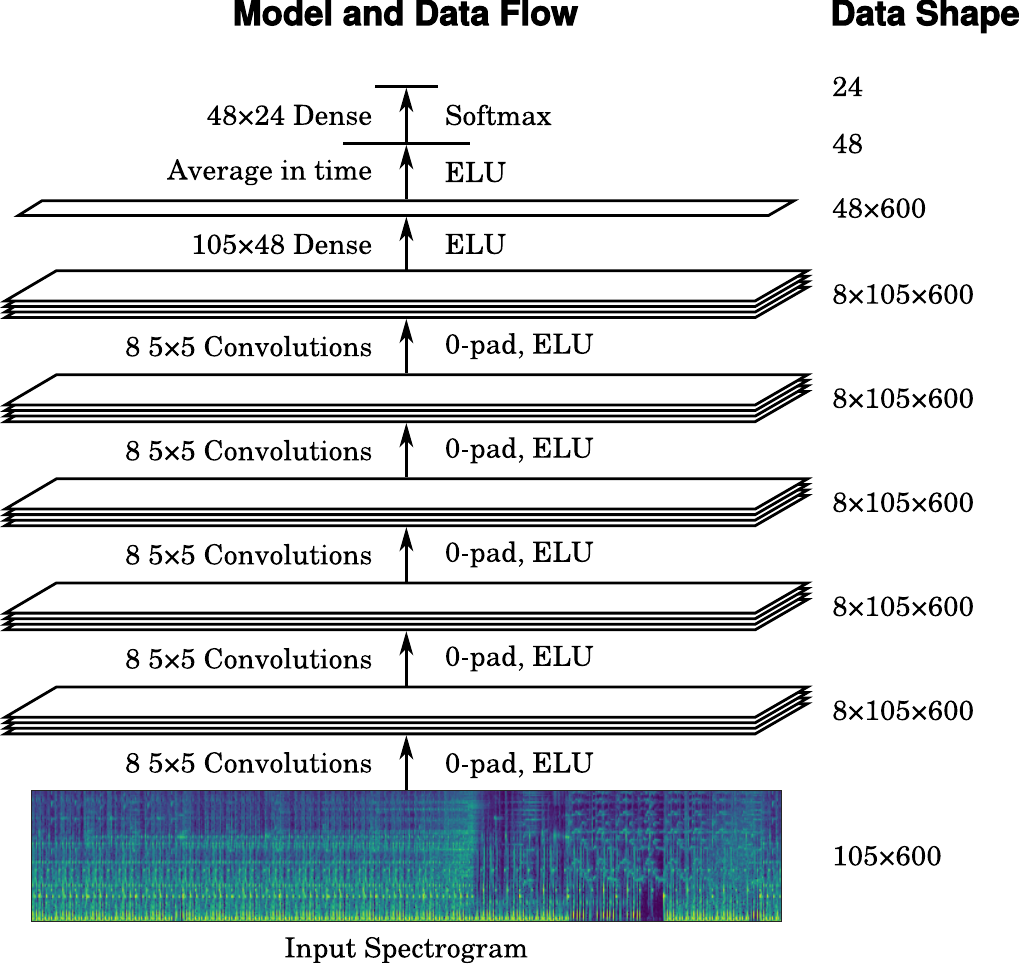}
\caption{Neural network for musical key estimation. Convolutions are 0-padded,
         i.e. the feature maps keep the input shape. All layers are followed
         by exponential-linear activations \cite{clevert_fast_2016}, except
         the last, which is followed by a softmax.}
\label{fig:model}
\end{figure}

The proposed neural network is designed to encompass all stages of the classic
key estimation pipeline: a pre-processing stage of convolutional layers, a
dense layer that projects the feature maps into a short representation at the
time-frame level, a global averaging layer that aggregates this representation
over time, and a softmax classification layer that predicts the global key of a
piece. Figure~\ref{fig:model} shows our model's architecture: five convolutional
layers with 8 feature maps computed by $\text{5}\times\text{5}$ kernels,
followed by a dense layer with 48 units applied frame-wise; this projection
is then averaged over time and classified using a 24-way softmax layer.
All layers (except the softmax layer) use the exponential-linear activation
function \cite{clevert_fast_2016}.

The convolutional layers constitute the first part of the ``feature
extraction'' equivalent in traditional key estimation systems. They are
intended to process the input spectrogram, deal with detrimental factors such
as noise or slight detuning, and, together with the projection layer, compute a
short frame-wise description of harmonic content. This part of the network can
process inputs of arbitrary lengths. Its output is aggregated in the following
layers.


Before classification, an averaging layer reduces the extracted representation
to a fixed-length vector. We could employ other, more powerful methods (like
recurrent layers), but we found in preliminary experiments that they fail to
achieve better results. 

Finally, a softmax classification layer predicts the global key for the audio.
We restrict ourselves to major and minor modes only, resulting in 24 possible
classes ($12\:\text{tonics}\times\{\text{major}, \text{minor}\}$)
as output. This is a common restriction, since most musical pieces are in
either major or minor, and as of now, there are no datasets with reliable
song-level annotations of other modes.

\subsection{Training}

We train the model using stochastic gradient descent with momentum,
back-propagating through the network the categorical cross-entropy error
between true key label $y_i$ and network output $\hat{y}_i$, and apply weight
decay with a factor of $10^{-4}$ for regularisation. The initial learning
rate is 0.001, with a momentum factor of 0.9. If validation accuracy did
not increase within 10 epochs, we halve the learning rate and continue training
with the parameters that gave the best results until then. After 100 epochs, we
select the model that achieved the best validation accuracy.

\section{Experiments}

Our experiments aim at 1) comparing the proposed system to reference systems,
and 2) examining the effect that the type of training data (in our case, the musical
genre) has on the results. Template-based algorithms require specialised key
templates for genres like electronic dance music in order to perform well
\cite{faraldo_key_2016}. We want to see if, and how strongly, our system is
affected by this. In the following, we discuss the data, the
evaluation metrics, the reference systems, and the different set-ups of our
system that we used in the experiments.

\subsection{Data}

We use three datasets in the course of our experiments: the GiantSteps key
dataset \cite{knees_two_2015}, the GiantSteps-MTG key dataset, and a subset of
the McGill Billboard dataset \cite{burgoyne_expert_2011}.

\newcommand{\GS}{GS}
The \textbf{GiantSteps Key
Dataset}\footnote{\url{https://github.com/GiantSteps/giantsteps-key-dataset}}
comprises 604 two-minute audio previews from \url{www.beatport.com}, with key
ground truth for each excerpt. It consists of various sub-genres of electronic
music. We use this dataset for testing purposes only, and will refer to it as \emph{\GS{}}.

\newcommand{\GSMTG}{GS\textsuperscript{MTG}}
The \textbf{GiantSteps MTG Key
dataset}\footnote{\url{https://github.com/GiantSteps/giantsteps-mtg-key-dataset}},
collected by \'Angel Faraldo from the Music Technology Group at Universitat
Pompeu Fabra, comprises 1486 two-minute audio previews from the same source.
These excerpts are distinct from the ones in the \GS{} dataset. From this
set, we only use excerpts labelled with a single key and a high confidence
for training (1077 pieces). We will refer to this dataset as \emph{\GSMTG{}}.

\newcommand{\BBTV}{BB\textsuperscript{TV}}
\newcommand{\BBTE}{BB\textsuperscript{TE}}
The \textbf{McGill Billboard
Dataset}\footnote{\url{http://ddmal.music.mcgill.ca/research/billboard}}
comprises 742 unique songs sampled from the American Billboard charts between
1958 and 1991, and thus consists of mostly pop and rock music. Unfortunately,
only the tonic, and not the mode (major or minor), is annotated for each piece.
We therefore estimate the mode using the tonic and chord annotations, following
a simple procedure for each piece: 1) select all chords whose root is the
tonic; 2) if more than 90\% of these chords are major, the key mode is assumed
to be major, and vice-versa for minor; 3) else, discard the song, because we
cannot confidently estimate the mode. Similarly, we discard songs with multiple
annotated tonics. This leaves us with 625 songs with key annotations. We
then divide the set into subsets of 62.5\% for training, 12.5\% for validation,
and 25\% for testing.  The exact division and key ground truths are available
online\footnote{\url{http://www.cp.jku.at/people/korzeniowski/bb.zip}}. 
We will refer to this dataset as \emph{\BBTV{}} and \emph{\BBTE{}} for the
train/validation and test sub-sets, respectively.

\subsection{Data Augmentation}

The datasets provide only few training data (1077 in \GSMTG{}, 391 in \BBTV{}),
compared to datasets used in computer vision (e.g.\ 40000 in CIFAR-10 or 60000
in MNIST). The generalisation capability of deep neural networks, however,
depends on a large number of training samples. We thus have to rely on data
augmentation to increase the number of training examples artificially.

Several augmentation techniques for audio input have been explored
\cite{schluter_exploring_2015}, with pitch shifting being particularly popular
in harmony-related tasks \cite{korzeniowski_fully_2016,cho_improved_2014,
humphrey_rethinking_2012}. Since pitch shifting is an expensive time-domain
operation, most works manipulate the time-frequency representation to emulate
it. In this work, however, we found that using a time-domain pitch shifting
algorithm\footnote{We used the SoX software available at \url{http://sox.sourceforge.net/}.}
directly on the audio gave better results in terms of classification accuracy.
We therefore shift each training song in the range of -4 to +7 semitones (and
adjust the target key accordingly), which increases the amount of training data
by a factor of 12.

\subsection{Metrics}

Evaluating key estimation results requires a more detailed quantitative
analysis than computing accuracy scores. In particular, although we consider
the task to be a simple 24-way classification problem when designing the
system, some classes are semantically closer to each other than others. For
example, the key of A-minor is called the ``relative minor'' to the key of
C-major, as they share all pitch classes, and differ only in the tonic.
Therefore, it is reasonable to consider some errors to be more severe than others.

The MIREX evaluation campaign\footnote{\url{http://www.music-ir.org/mirex}}
developed an evaluation strategy and introduced a single weighted measure that
reflects the above considerations. Following their guidelines, key predictions
can fall into the following categories:

\begin{itemize}
        \item \textbf{Correct:} if the tonic and the mode (major/minor) of
                prediction and target correspond.
        \item \textbf{Fifth:} if the tonic of the prediction is the fifth of
                the target (or vice versa), and modes correspond.
        \item \textbf{Relative Minor/Major:} if modes differ and either 
                a) the predicted mode is minor and the predicted tonic is 3 
                semitones below the target, or
                b) the predicted mode is major and the predicted tonic is 3
                semitones above the target.
        \item \textbf{Parallel Minor/Major:} if modes differ but the predicted
                tonic matches the target.
        \item \textbf{Other:} Prediction errors not caught by any category, 
                i.e.\ the most severe errors.
\end{itemize}

We first compute the ratio of predictions that fall into each category. We then
calculate the MIREX \textbf{weighted} score as \linebreak $w = r_c + 0.5 \cdot r_f + 0.3
\cdot r_r + 0.2 \cdot r_p$, where $r_c$, $r_f$, $r_r$, and $r_p$ are the ratios
of the correct, fifth, relative minor/major, and parallel minor/major,
respectively. These ratios reveal more about the capability of the algorithms
than accuracy (i.e., the ``correct'' ratio) alone. They allow us to see the
kind of mistakes the system makes, and at the same time, assign a single number
for comparing its performance with others. 

The ratios of the individual error categories cannot be compared in isolation,
but only in context with the other ratios. The only numbers that can be
compared individually are the weighted score (because it aggregates all error
types), percentage of correct classifications (because it corresponds to
classification accuracy), and the ``other'' error ratio (because it tells us
how often a system predicts unrelated keys). That is why, in
Table~\ref{tab:results}, we will highlight the best results only for these
categories.

\subsection{Setups and Reference Systems}

\newcommand{\CK}[1]{CK\textsuperscript{#1}}
\newcommand{\EDM}[1]{EDM\textsuperscript{#1}}
\begin{table*}[]
\centering
\caption{Results of various training configurations of our proposed model and
        of reference systems. Boldface indicates best results. \GS{} and \GSMTG
        refer to the GiantSteps datasets (electronic music),
        BB\textsuperscript{*} to various subsets of the Billboard dataset
        (pop/rock music). \CK{*} denote the proposed model trained on
        different data sets, \EDM{*} and QM denote 
reference systems by \cite{faraldo_key_2016,cannam_mirex_2016}.}
\label{tab:results}
\begin{tabular}{@{}lllrrrrrr@{}}
\toprule
\textbf{Test Set} & \textbf{Method} & \textbf{Train Set} & \textbf{Weighted} & \textbf{Correct} & \textbf{Fifth} & \textbf{Relative} & \textbf{Parallel} & \textbf{Other} \\ \midrule
\GS               & \CK{1}          & \GSMTG             & \textbf{74.3}     & \textbf{67.9}    & 6.8            & 7.1               & 4.3               & \textbf{13.9}  \\
                  & \CK{2}          & \BBTV              & 57.3              & 47.0             & 6.5            & 12.6              & 16.6              & 17.4           \\
                  & \CK{3}          & \GSMTG, \BBTV      & 69.2              & 61.9             & 6.8            & 8.6               & 6.3               & 16.4           \\
                  & \EDM{A}            &                    & 65.6              & 57.8             & 7.3            & 6.6               & 10.8              & 17.6           \\
                  & \EDM{M}            &                    & 70.1              & 63.7             & 8.6            & 2.7               & 6.5               & 18.5           \\
                  & \EDM{T}            &                    & 44.6              & 33.6             & 8.8            & 15.4              & 9.9               & 32.3           \\
                  & QM              &                    & 50.4              & 39.6             & 11.9           & 13.2              & 4.3               & 31.0           \\ \midrule
\BBTE             & \CK{1}          & \GSMTG             & 72.8              & 62.5             & 7.6            & 13.2              & 12.5              & \textbf{4.2}   \\
                  & \CK{2}          & \BBTV              & \textbf{83.9}     & \textbf{77.1}    & 9.0            & 4.9               & 4.2               & 4.9            \\
                  & \CK{3}          & \GSMTG, \BBTV      & 79.7              & 70.8             & 9.7            & 9.0               & 6.3               & \textbf{4.2}   \\
                  & \EDM{A}            &                    & 78.7              & 70.8             & 11.8           & 2.8               & 5.6               & 9.0            \\
                  & \EDM{M}            &                    & 28.9              & 14.6             & 2.1            & 16.0              & 42.4              & 25.0           \\
                  & \EDM{T}            &                    & 75.4              & 66.7             & 12.5           & 6.3               & 2.8               & 11.8           \\
                  & QM              &                    & 60.9              & 52.1             & 11.8           & 4.2               & 8.3               & 23.6           \\ \bottomrule
\end{tabular}
\end{table*}

We train our method in three configurations: \CK{1}, trained on \GSMTG{};
\CK{2}, trained on \BBTV{}; and \CK{3}, trained on both \GSMTG{} and \BBTV{}.
We evaluate each of the trained models on \GS{} and on \BBTE{}. This way, we
observe the system's performance when trained on the \emph{same} genres as it
is tested on, when trained on a \emph{different} genre (the cross-genre setup),
and when trained on multiple genres (to see if it can learn a unified model for
different genres).

We compare our method against the Queen Mary Key Detector (QM) \cite{cannam_mirex_2016} and three
variations of the method presented in \cite{faraldo_key_2016} (\EDM{A}, \EDM{M}, \EDM{T}).
For both systems, open source implementations are available\footnote{\url{http://vamp-plugins.org/plugin-doc/qm-vamp-plugins.html}\\
\url{https://github.com/angelfaraldo/edmkey}}.
QM consists of a hand-crafted pre-processing stage and correlates the
obtained chromagrams with key profiles based on Bach's Well Tempered Klavier.
The three EDM systems also use a hand-crafted pre-processing stage, but use
different key profiles for classification: \EDM{A} uses key profiles
automatically derived from a set of electronic music; \EDM{M} uses hand-tuned
profiles based on the automatically derived ones (effectively disabling the
prediction of major keys); \EDM{T} uses profiles based on European classical
music. Of all submissions to MIREX, \EDM{M} and QM achieved the best results on
the electronic and classical music datasets used for evaluation, respectively.
We thus consider both to be state of the art.

\subsection{Results}

Table~\ref{tab:results} shows the evaluation results of all training
configurations of our proposed model, and of the reference systems. We
determine the statistical significance of the results using a Wilcoxon signed
rank test, with the error types representing the ranks.  If trained on the
correct genre, our model clearly outperforms the reference systems: 74.3 vs.
70.1 ($\alpha = 0.001$) for the GiantSteps dataset,  and 83.9 vs. 78.7 ($\alpha
= 0.014$) for the Billboard dataset.

Examining the cross-genre setups, we observe a significant drop in key
estimation accuracy: tested on \GS{} (electronic music), a model trained on
\BBTV{} (pop/rock) achieves a weighted score of only 57.3, compared to 74.3
when trained on electronic music from \GSMTG{}. However, we also see that the
number of severe mistakes (category ``other'') that our system commits in this
setup is not higher than those of the reference systems: the model only
predicts a completely unrelated key 17.4\% of the time---similar to the
reference systems specialised on this genre (17.6\% and 18.5\% for \EDM{A} and
\EDM{M}, respectively); vice-versa, when trained on \GSMTG{} and evaluated on
\BBTE{}, it achieves the lowest rate of severe mistakes (4.2\%).

The most common error occurring in these cross-genre setups is predicting the
wrong mode (resulting in parallel minor/major) and predicting the relative
minor/major key. This suggests that while the model is still able recognise
some fundamental concepts of tonality, finer characteristics vary too much
between pieces of different genres.

We wanted to see if the proposed model can be trained to provide a good unified
key estimator for multiple genres by combining training data. The resulting
system \CK{3} does not reach the performance of the specialised ones (69.2 vs
74.3 on \GS{}, 79.7 vs. 83.9 on \BBTE{}); however, on \GS{}, it performs as
well as \EDM{M}, which is tuned manually to give good results on electronic
music datasets (69.2 vs. 70.1, $\alpha = 0.94$). It still performs better
than \EDM{A}, which is also trained on electronic music, but without manual
post-training adaptations (69.2 vs. 65.5, $\alpha = 0.02$).

The numbers presented for the \EDM{*} systems for the GiantSteps dataset differ
from the ones originally reported in \cite{faraldo_key_2016}. This is
mainly because we applied a stricter criterion for the ``fifth'' category: we
require the predicted mode to match the target mode, while
\cite{faraldo_key_2016} ignores the mode for this category. Also, according to
personal correspondence with the author, changes in the library used in the
original implementation worsened the results relative to the original ones.

\section{Conclusion and Future Work}

We have presented a global key estimation system based on a convolutional
neural network. Compared to previous work, this model can be automatically
trained end-to-end, without the need of expert knowledge in feature design or
specific pre-processing steps such as tuning correction or spectral whitening.

We have shown experimentally that the model performs state-of-the-art on
datasets of electronic music and pop/rock music. Additionally, we are planning
to evaluate the proposed model on more genres, e.g.\ classical music. 

Another feature of the proposed model is its ability to adapt to multiple types
of music without changing the model itself; it just needs to be re-trained with
a training set extended to the type of music of interest. While it still showed
good performance in such a scenario, it did not reach the level of its
specialised counterparts.

A clear limitation of the proposed method is that it only estimates a global
key for a complete piece. While this is adequate for certain types of music,
other types (e.g.\ classical music) involve key modulations that our method
currently cannot capture. A possible easy fix could be to apply our model using
a sliding window over the spectrogram. Extending the proposed method in such a
way is left to future work.

Finally, we have to keep in mind that even with data augmentation, we are still
working with small datasets. Although we increase the number of training
samples by a factor of 12 using pitch-shifting, this is not equivalent to
having available 12 times as many musical pieces: the musical content of the
artificial data points is still the same as in the seed data point, just in a
different key. We expect the system's performance to improve once more training
data is available.

\section*{Acknowledgments}

This work is supported by the European Research Council (ERC) under the EU’s
Horizon 2020 Framework Programme (ERC Grant Agreement number 670035, project
``Con Espressione''). The Tesla K40 used for this research was donated by the
NVIDIA Corporation.

\balance

\bibliographystyle{IEEEtran}
\bibliography{IEEEabrv,eusipco2017}

\end{document}